\documentclass[11pt]{article}
\pdfoutput=1

\usepackage[a4paper,margin=1in]{geometry}
\usepackage{mathtools}
\usepackage{graphicx}
\usepackage{url} % not crucial - just used below for the URL 
\usepackage{float}
\usepackage{setspace, bm} 
\usepackage{enumerate, enumitem}
\usepackage{subcaption} 
\usepackage{threeparttable}

\usepackage{multirow}
\usepackage{verbatim}
\usepackage{amssymb}
\usepackage[table]{xcolor}
\usepackage{booktabs}
\usepackage{changepage}
\usepackage[T1]{fontenc}

\usepackage{rotating} % add this in the preamble

\usepackage{mathtools}
\usepackage{graphicx}
\usepackage{url} % not crucial - just used below for the URL 
\usepackage{float}
\usepackage{setspace, bm} 

\usepackage{enumerate, enumitem}

\usepackage{rotating} % add this in the preamble

\usepackage{algorithm}
\usepackage{algorithmic}
\def\bSig\mathbf{\Sigma}

\usepackage{amssymb, amsmath}

\usepackage{siunitx}
\PassOptionsToPackage{hyphens}{url}\usepackage[hidelinks]{hyperref}
\usepackage{cleveref}
\usepackage[utf8]{inputenc}
\usepackage[right]{lineno}
\usepackage{csquotes}
\usepackage{booktabs}
\usepackage{longtable}
\usepackage{adjustbox}
\usepackage{array}
\usepackage{titlesec}
\usepackage{authblk}
\usepackage{xcolor} % Load the xcolor package for color options

\usepackage{listings}
\usepackage{nicefrac}       % compact symbols for 1/2, etc.
\usepackage{microtype}      % microtypography
\usepackage{graphicx}
\usepackage{natbib}
\usepackage{doi}

\usepackage[english]{babel}

\title{Diagnosis-based mortality prediction for intensive care unit patients via transfer learning}
\author{Mengqi Xu\thanks{m332xu@uwaterloo.ca}}
\author{Subha Maity\thanks{subha.maity@uwaterloo.ca}}
\author{Joel Dubin\thanks{jdubin@uwaterloo.ca}}

\affil{Department of Statistics and Actuarial Science \\
University of Waterloo}

\date{}

\def\bSig\mathbf{\Sigma}

\begin{document}
{\singlespacing
\maketitle}
\vspace*{-3em}

\begin{abstract}
    In the intensive care unit, the underlying causes of critical illness vary substantially across diagnoses, yet prediction models accounting for diagnostic heterogeneity have not been systematically studied. To address the gap, we evaluate transfer learning approaches for diagnosis-specific mortality prediction and apply both GLM- and XGBoost-based models to the eICU Collaborative Research Database. Our results demonstrate that transfer learning consistently outperforms models trained only on diagnosis-specific data and those using a well-known ICU severity-of-illness score, i.e., APACHE IVa, alone, while also achieving better calibration than models trained on the pooled data. Our findings also suggest that the Youden cutoff is a more appropriate decision threshold than the conventional 0.5 for binary outcomes, and that transfer learning maintains consistently high predictive performance across various cutoff criteria. 

\end{abstract}

\section{Introduction}
\label{sec:introduction}
% use APACHE score to predict
% demonstrate XGBoost outstands in eICU-CRD
% have research on specific single diagnosis
% gap in per diagnosis
% transfer learning canbe used for such kinda problem
% gap of TL in eICU-CRD
% we investigate TL in eICU to see their performance (improvement or not)
Accurate prediction of in-hospital mortality for intensive care unit (ICU) patients is crucial for clinical decision-making and patient management, as these patients often present with severe, life-threatening conditions requiring continuous monitoring and timely intervention \citep{becker2019telemedicine}. 
For example, in the eICU Collaborative Research Database (eICU-CRD) \citep{pollard2018eicu}, patients are admitted with a wide range of acute and unpredictable conditions, and the underlying causes of life-threatening illnesses often differ substantially across diagnoses. Consequently, developing reliable predictive models for in-hospital mortality that are tailored to each diagnosis is crucial for a more accurate and comprehensive prediction.

Many recent research studies have focused on predicting in-hospital mortality among ICU patients. The Acute Physiology and Chronic Health Evaluation (APACHE) score, for example, is a widely used prognostic tool for estimating in-hospital mortality in critical care settings \citep{zimmerman2006acute}; the currently improved version of this score is now APACHE IVa. Despite its broad application and overall usefulness for predicting in-hospital mortality and length of stay of ICU patients, APACHE can sometimes produce false positives and false negatives \citep{feng2021identifying}. 
% identified several early-measured variables associated with inaccurate APACHE IVa predictions of in-hospital mortality in eICU-CRD.
Beyond traditional scoring systems, machine learning approaches have increasingly been applied to mortality prediction. \citet{zhao2023improving}, for instance, compared multiple models and found that XGBoost achieved excellent discrimination when trained across all diagnostic groups in eICU-CRD. However, the performance of such global models within individual diagnostic groups remains insufficiently understood.
Although other studies have developed explainable models tailored to specific diagnoses, such as acute pancreatitis \citep{ren2024prediction} and congestive heart failure \citep{han2022early}, no prior study has systematically evaluated in-hospital mortality prediction across the full spectrum of diagnostic categories in ICU patients while accounting for diagnosis-specific heterogeneity. This gap limits the understanding of diagnosis-level model performance and constrains the generalizability of existing predictive approaches.

To address this gap, we investigate prediction models for in-hospital mortality of ICU patients stratified by diagnosis, demonstrating our methodology on the eICU-CRD. A key challenge arises from the substantial variation in sample sizes across diagnostic categories. While some diagnoses, such as sepsis or cardiac arrest, are well represented, many others, including trauma and gastrointestinal (GI) obstruction, have far fewer observations; see \autoref{tab:diagnosis_summary} for patient counts per diagnosis. Modeling each diagnosis group independently is not a viable solution, as it may lead to unstable estimates, especially in small-sample settings. Pooling all patients into a single model, on the contrary, may obscure diagnosis-specific patterns and introduce bias due to the substantial differences in underlying disease mechanisms across diagnoses.

Transfer learning offers a promising solution to this problem. As a set of methods designed to improve performance in a target domain with limited data by leveraging information from related source domains, transfer learning allows models to borrow information from data-rich diagnostic groups while preserving diagnosis-specific characteristics. 
In clinical research, this approach has been developed and applied in diverse contexts. For example, \citet{maity2024linear} proposed a linear adjustment–based transfer learning method and applied it to mortality prediction among British Asians using UK Biobank data \citep{sudlow2015uk}.
\citet{guha2025enhancing} developed a transfer learning framework to enhance causal inference for small cohorts and applied it to lung sepsis patients in the eICU-CRD. \citet{wang2025transfer} explored a transfer learning algorithm for precision medicine and applied it to sepsis patients from eICU-CRD and MIMIC-III \citep{johnson2016mimic}.
\citet{desautels2017using} applied transfer learning to improve mortality prediction in a data-scarce hospital setting. 
However, to our knowledge, transfer learning has not been systematically investigated for diagnosis-specific mortality prediction in ICUs. Confronting this challenge is a critical step toward more accurate and comprehensive models that can produce an improved diagnosis-specific predictions.

In this study, we evaluate transfer learning methods for in-hospital mortality prediction across diagnostic categories in the eICU-CRD. 
We compare their performance against source models, target-only models, as well as models using only the APACHE IVa score. 
The remainder of this paper is organized as follows.
\autoref{Data} describes eICU-CRD and the cohort selection process.
\autoref{Method} presents the transfer learning methods for GLM and XGBoost \citep{chen2016xgboost}, along with other models considered in this study.
\autoref{ApplicationResult} reports the results of the data application on eICU-CRD, comparing discrimination, calibration, and other predictive accuracy measures between transfer learning models and alternative approaches.
\autoref{simulation} details a simulation study designed to further examine the ceiling effect of model discrimination measures, namely the area under the receiver operating characteristic curve (AUROC) and the area under the precision-recall curve (AUPRC), observed in the analysis of the eICU-CRD dataset presented in \autoref{ApplicationResult}. This study provides a more comprehensive assessment of the performance of the proposed transfer learning methods under varying conditions.

%% Additional sections
\section{Data}\label{Data}
% add flowchart (R)
\subsection{Data description and cohort selection}\label{DataDescription}
\begin{figure}[h]
    \centering
    \includegraphics[width=0.8\linewidth]{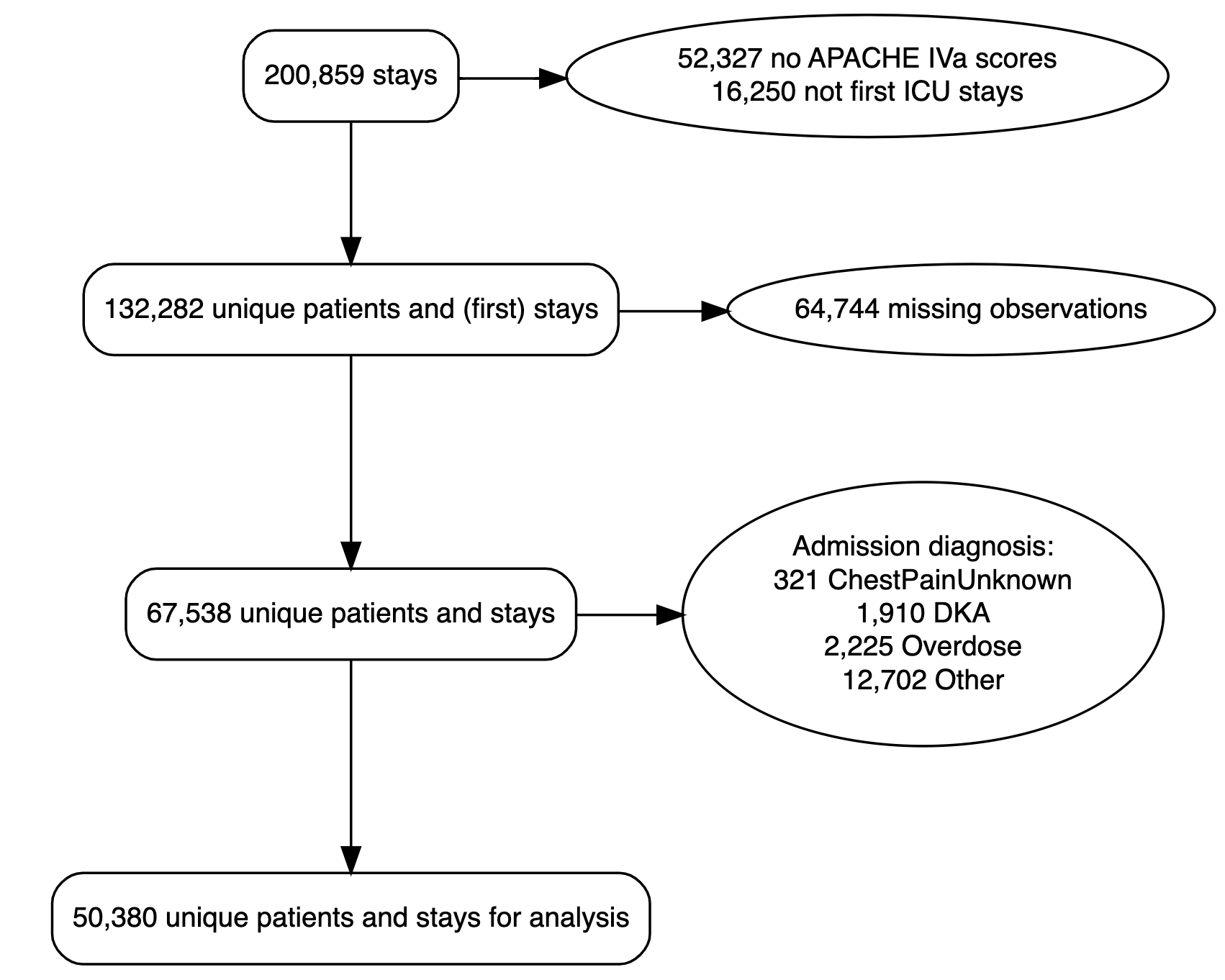}
    \caption{Flowchart of cohort selection from eICU-CRD and data preparation process.}
    \label{fig:flowchart}
\end{figure}

The eICU Collaborative Research Database (version 2.0) \citep{pollard2018eicu} contains de-identified clinical data from over 200,000 ICU admissions across more than 200 hospitals in the United States between 2014 and 2015. We restrict our analysis to patients with documented APACHE IVa scores. To ensure one observation per patient and enable early outcome prediction, we select the first hospital admission per patient and then the first unit (i.e., ICU) stay within that hospital stay. \autoref{fig:flowchart} illustrates our cohort selection and data preparation process. %Patients with missing gender labels were excluded to minimize missingness.

The primary binary outcome of interest is in-hospital mortality, defined as death during the ICU stay or prior to hospital discharge.
We want to predict mortality using a range of patient-level variables, covering demographics, laboratory measurements, vital signs, and prior medical history. Hospital-level information was also incorporated via the teaching status of the admitting institution.
Demographic covariates included age, ethnicity, and binary gender. 
% Information regarding patient length of stay—both in the ICU and in the hospital overall—was also recorded.
Laboratory variables were summarized by computing the frequency of lab test measurements taken during the first 24 hours of the ICU stay. These tests included pH, bicarbonate, lactate, creatinine, and other standard chemistry values. 
% pH, fibrinogen, magnesium, calcium, chloride, phosphate, bicarbonate, amylase, lactate, lipase, creatinine
The rationale for using lab test counts—as opposed to only test values—is to partially capture the clinical intensity and acuity of care, which reflects underlying severity.
Vital sign data were derived from time-series measurements taken within the first 24 hours of ICU admission. For each patient, we extracted functional principal component (FPCs) scores \citep{benko2009common} from the heart rate, respiratory rate, and oxygen saturation (SaO$_2$) trajectories to capture key patterns of physiological evolution in a compact form, with the threshold of variation explained by the leading FPCs set at 95\%.

We also included a set of clinically derived severity features, referred to as the APACHE Acute Physiology Score (APS) variables, which combine the worst vital signs and laboratory values recorded during the first 24 hours of each patient’s ICU stay. These variables contribute to the APS calculation and provide a concise summary of the patient's most extreme physiological status at the time of ICU admission.
In addition, we included patients' admission weight and height as markers of body composition status. We also incorporated relevant medical history, including pre-operative myocardial infarction, cardiac catheterization, and percutaneous transluminal coronary angioplasty. All records with missing values in any of the selected variables were excluded from the analysis; handling missing data will be a focus of future research. This resulted in a remaining cohort of 67,538 unique patient ICU stays.

\subsection{Admission diagnosis groups (target data)}
Admission diagnoses were provided in text form in the database; we mapped them into 21 clinically coherent categories based on \citet{cosgriff2019developing}, as displayed in \autoref{tab:diagnosis_summary}. The table also provides summaries of patient numbers and mortality outcomes by each diagnosis.
% , i.e., acute coronary syndrome (ACS), acute renal failure (ARF), asthma emphysema, coronary artery bypass grafting (CABG), cardiac arrest, chest pain, congestive heart failure (CHF), coma, cerebrovascular accident/stroke (CVA), diabetic ketoacidosis (DKA), gastrointestinal (GI) bleed, GI obstruction, neurologic problems, overdose, pneumonia (PNA), sepsis, trauma, valve repair/replacement, other cerebrovascular problems, other respiratory problems, and other diagnoses. 
Although the original classification includes 21 diagnosis categories, we exclude the classification ``Other'' due to 
% the diagnosis category being the focus of our methodology and
the ambiguity of the clinical meaning of this specific diagnosis. We also exclude diagnoses with too few observations (< 20) for in-hospital mortality, such as ``ChestPainUnknown'', ``DKA'', and ``Overdose''.
% “Other”, “ChestPainUnknown”, "DKA", "Overdose" from our modeling due to ambiguous clinical meanings and limited observations (under 20) of death. 
The final analysis includes 17 diagnosis groups.

\begin{table}[htbp]
\centering
\caption{Total patient counts, counts of in-hospital mortality, and prevalence (mortality rate) per diagnostic group in eICU-CRD.}
\label{tab:diagnosis_summary}
\begin{tabular}{lrrrr}
\toprule
\textbf{Diagnosis Group} & \textbf{Expired} & \textbf{Alive} & \textbf{Total} & \textbf{Prevalence (\%)} \\
\midrule
GIObstruction     & 71 & 525 & 596 & 11.91 \\
ARF               & 75 & 879 & 954 & 7.86 \\
Coma              & 94 & 957 & 1051 & 8.94 \\
ValveDz           & 31 & 1327 & 1358 & 2.28 \\
CVOther           & 71 & 1654 & 1725 & 4.12 \\
Asthma--Emphys    & 122 & 1879 & 2001 & 6.10 \\
PNA               & 318 & 1996 & 2314 & 13.74 \\
CHF               & 263 & 2245 & 2508 & 10.49 \\
Neuro             & 71 & 2441 & 2512 & 2.83 \\
CABG              & 43 & 2508 & 2551 & 1.69 \\
Trauma            & 211 & 2899 & 3110 & 6.78 \\
GIBleed           & 244 & 3468 & 3712 & 6.57 \\
RespMedOther      & 523 & 3519 & 4042 & 12.94 \\
ACS               & 176 & 3996 & 4172 & 4.22 \\
CardiacArrest     & 973 & 3429 & 4402 & 22.10 \\
CVA               & 514 & 4346 & 4860 & 10.58 \\
Sepsis            & 1363 & 7149 & 8512 & 16.01 \\
\midrule
Total & 5163 & 45217 & 50380 & 10.25 \\
\toprule
\textbf{Removed from analysis} & & & &  \\
ChestPainUnknown & 5 & 316 & 321 & 1.56 \\
DKA               & 14 & 1896 & 1910 & 0.73 \\
Overdose          & 17 & 2208 & 2225 & 0.76 \\
Other             & 688 & 12014 & 12702 & 5.42 \\
\midrule
Total & 5887 & 61651 & 67538 & 8.72 \\
\bottomrule
\end{tabular}
\end{table}

\section{Method}\label{Method}
We evenly split each target data subset (i.e., diagnosis group) into three folds using stratified cross-validation, ensuring that each fold contains both alive and deceased patients. This approach addresses the highly imbalanced outcome distribution and improves the robustness of model evaluation. Within each fold, we train target-only models and transfer learning models on the training fold, while source models are trained on the entire dataset, excluding the test fold.

Our objective is to predict in-hospital mortality for patients within each diagnostic group. 
% \autoref{tab:diagnosis_summary} summarizes the number of patients and mortality outcomes across diagnostic groups. 
% From \autoref{tab:diagnosis_summary} recall that there is considerable variability in group sizes, ranging from around 500 patients (e.g., GIObstruction) to over 6000 (e.g., Sepsis), and in-hospital mortality is highly imbalanced within each group, with a relatively low percentage of patients experiencing the event of interest. These characteristics pose challenges for building well-calibrated, robust models using only diagnosis-specific data.
We apply both regression-based and XGBoost-based transfer learning methods to enhance mortality prediction within each diagnosis group. In our context, each diagnosis group is treated as a separate target task, while the overall patient population serves as the source. By borrowing strength across groups, the transfer learning approach enables more stable estimation and improved predictive accuracy, particularly in small diagnostic cohorts.
In the following subsections, we introduce the transfer learning algorithms used in our study and describe their implementations on eICU-CRD.

\subsection{Notations}
% Each patient is represented by 51 baseline variables, which are detailed in \autoref{DataDescription}, including demographics, laboratory results, vital signs, prior history, and severity scores, along with all pairwise interaction terms.

We consider \( K ( = 17)\) target datasets, each corresponding to a different diagnosis group. Let \( m_k \) denote the sample size of the \( k \)-th target dataset, for \( k = 1, \dots, K \). In each target dataset, we observe data \( (\boldsymbol{x}_{k,i},y_{k, i}) \in \mathbb{R}^p \times \{0, 1\} \) for $i = 1, ..., m_k$, where \( \boldsymbol{x}_{k,i} \in \mathbb{R}^p \) is the feature vector for the \( i \)-th patient, and \( y_{k, i} \in \{0, 1\} \) is the binary indicator of in-hospital mortality (\( y_{k, i} = 1 \) if the patient expired).
The source dataset is constructed by pooling all target datasets, yielding a total sample size of \( n = \sum_{k=1}^K m_k \).

Given that the response outcome is binary, we define the loss function for each observation $i$ as the following negative log-likelihood:  
\[
\ell(y_{k, i}, \hat{\eta}_{k, i}) = - \left[ y_{k, i} \log \sigma(\hat{\eta}_{k, i}) + (1 - y_{k, i}) \log(1 - \sigma(\hat{\eta}_{k, i})) \right],
\]
where $\hat{\eta}$, as a function of $\boldsymbol{x}$, denotes the log-odds of the predicted probability of in-hospital death, and \(\sigma(\cdot)\) is the inverse of the logistic link function, commonly known as the sigmoid function, and is defined as $\sigma(t) = \{1 + e^{-t}\}^{-1}, ~ t \in \mathbb{R}$.
For simplicity, we define the collection of all two-way interactions and quadratic terms as $\boldsymbol{x} \otimes \boldsymbol{x}$, and  we define the covariate transformation function as 
\[
\phi(\boldsymbol{x}) = \begin{cases}
\boldsymbol{x}, & \text{(main effects only)} \\
(\boldsymbol{x}, \boldsymbol{x} \otimes \boldsymbol{x}), & \text{(main and interaction effects)}
\end{cases}
\]
% where \(x_i \otimes x_i\) denotes the element-wise interactions between features. 
% Let \(\delta\) be the adjustment vector corresponding to \(\phi(x_i)\). 

\subsection{Two-step transfer learning}
To indicate the fitted model parameter values on the source (pooled) datasets, we use the subscript \(s\).

\subsubsection{GLM-based transfer learning}\label{GLM-based transfer learning}

For GLM-based transfer learning, regularization is performed in both Steps 1 and 2.

\noindent\textbf{Step 1:} Estimate a base model on the pooled dataset
\[
\hat{\boldsymbol{\alpha}}_s, \hat{\boldsymbol{\beta}}_s = 
\arg \min_{\boldsymbol{\alpha}_s, \boldsymbol{\beta}_s} 
\sum_{k = 1}^K\sum_{i=1}^{m_k} \ell\left( y_{k, i}, \boldsymbol{\alpha}_s^\top \boldsymbol{x}_{k,i} + \boldsymbol{\beta}_s^\top (\boldsymbol{x}_{k,i} \otimes \boldsymbol{x}_{k,i}) \right) 
+ \lambda_1 \, \lVert (\boldsymbol{\alpha}_s, \boldsymbol{\beta}_s) \rVert_1
\]

\noindent\textbf{Step 2:} Adjust the effect differences for each target cohort

\[
\hat{\boldsymbol{\delta}}_k = 
\arg \min_{\boldsymbol{\delta}} 
\sum_{i=1}^{m_k} \ell\left( y_{k, i}, \hat{\eta}_{k, i} + \boldsymbol{\delta}^\top \phi(\boldsymbol{x}_{k,i}) \right)
+ \lambda_{2, k} \lVert \boldsymbol{\delta} \rVert_1,
\]
where $\hat{\eta}_{k, i} = \hat{\boldsymbol{\alpha}}_s^\top \boldsymbol{x}_{k,i} + \hat{\boldsymbol{\beta}}_s^\top (\boldsymbol{x}_{k,i} \otimes \boldsymbol{x}_{k,i})$ and $\delta$ capture the target-specific effect deviation.
\\

\noindent\textbf{Final prediction:} For each test observation $\boldsymbol{x}_{k,i}$ with the $k$-th diagnosis, we predict the probability of in-hospital mortality as
\[
\text{logit}\left( \hat{p}_k(\boldsymbol{x}_{k,i}) \right) = \hat{\eta}_{k, i} + \hat{\boldsymbol{\delta}}_k^\top \phi(\boldsymbol{x}_{k,i}),
\quad\text{or equivalently,}\quad
\hat{p}_k(\boldsymbol{x}_{k,i}) = \sigma\left( \hat{\eta}_{k, i} + \hat{\boldsymbol{\delta}}_k^\top \phi(\boldsymbol{x}_{k,i}) \right).
\]
% where \(\sigma(t) = \{1 + e^{-t}\}^{-1}\) is the sigmoid function.

% \textbf{Step 3:} Estimate adjusted parameter for the target 
% (Or should we make joint estimation using source and target data?)
% $$\hat{\boldsymbol{\alpha}} = \hat{\boldsymbol{\alpha}}_p + \hat{\boldsymbol{\delta}}_\alpha, 
% \ 
% \hat{\boldsymbol{\beta}} = \hat{\boldsymbol{\beta}}_p + \hat{\boldsymbol{\delta}}_\beta,$$
% and use them to predict in-hospital mortality for each patient.

\subsubsection{XGBoost-based transfer learning}\label{XGBoost-based transfer learning}

\citet{zhao2023improving} showed that XGBoost had the best performance in predicting in-hospital mortality in the eICU-CRD among the candidate models investigated. Motivated by this finding, we implemented a two-step transfer learning procedure using XGBoost as the base model on the pooled data. This procedure shares similarities with the two-step structure of the GLM-based method described in \autoref{GLM-based transfer learning}. The only difference lies in the base model: we replace a regularized GLM with a tree-based XGBoost model to better capture nonlinearities and complex interactions in the data. 
Specifically:

\noindent \textbf{Step 1: Train the source model with XGBoost.} We fit an XGBoost classifier on the source dataset
% using the pooled dataset 
$\{(\boldsymbol{x}_{k,i}, y_{k, i}): k = 1, \dots, K; i = 1, \dots, m_k\}$. Denote the predicted log-odds on the target data as $\hat{\eta}_{k, i} = \text{XGB}(\boldsymbol{x}_{k,i})$, which will serve as a fixed offset term in Step 2.

\noindent \textbf{Step 2: Adjust target-specific effects via penalized logistic regression.} On the target data, we fit a logistic regression with an offset $\hat{\eta}_{k, i}$:

\begin{equation}
\hat{\boldsymbol{\delta}}_k = \arg\min_{\boldsymbol{\delta}} \sum_{i=1}^{m_k} \ell\left( y_{k, i}, \hat{\eta}_{k, i} + \boldsymbol{\delta}^\top \phi(\boldsymbol{x}_{k,i}) \right) + \lambda_k \left\| \boldsymbol{\delta} \right\|_1.
\end{equation}
where $\delta$ captures the target-specific effect deviation.

\noindent \textbf{Final prediction:} For each test observation $\boldsymbol{x}$ with $k$-th diagnosis, we predict the probability of in-hospital mortality via:
\[
\hat{p}_k(\boldsymbol{x}) = \sigma(\text{XGB}(\boldsymbol{x}) + \hat{\boldsymbol{\delta}}_k^\top \phi(\boldsymbol{x})).
\]

\subsection{Recalibration}\label{Recalibration}

To improve the calibration of predicted probabilities, we apply post-hoc recalibration within each diagnosis group \citep{steyerberg2019clinical}. Specifically, we perform a logistic recalibration procedure (also known as Platt scaling) to simultaneously address both over-/under-estimation of risk and overly extreme or conservative predictions.

We fit a logistic regression model where the predicted log-odds are entered as a covariate, allowing both the intercept and slope to be adjusted:
\[
\text{logit}\{P(y_{\text{new}} = 1\mid \boldsymbol{x})\} = a + b \cdot \text{logit}\left( \hat{p}(\boldsymbol{x}) \right),
\]
where $a$ is the recalibrated intercept, \( b \) is the estimated calibration slope, and $\hat p(\boldsymbol{x})$ is the target predictor from \autoref{GLM-based transfer learning} and \autoref{XGBoost-based transfer learning}. A slope of \( b = 1 \) and an intercept of $a = 0$ indicate perfect calibration; values of $b<1$ suggest overfitting (i.e., predictions that are too extreme), while values of $b>1$ greater than 1 suggest underfitting.
% Overall, our results show that the calibration slopes $b$ of XGBoost-based models are consistently greater than 1 and generally larger than those of GLMs, indicating a tendency toward underfitting.
With this recalibration, we ensure that both the spread and the central tendency of predicted probabilities are aligned with the observed outcomes. All recalibration steps are performed separately within each diagnostic group.

% We evaluate model performance using area under the receiver operating characteristic curve (AUROC), area under the precision-recall curve (AUPRC), Brier Score, and Integrated Calibration Index (ICI), to jointly assess discrimination and calibration quality across all models and diagnosis groups.

\subsection{Other models for comparison}

To benchmark the performance of the transfer learning approach, we compared it with several alternative models, including clinically established scoring systems and standard statistical learning approaches.

\paragraph{APACHE IVa-based baselines.}
Since the APACHE score is widely used to predict in-hospital mortality in the ICU \citep{zimmerman2006acute}, we used the most up-to-date version in eICU-CRD, i.e., APACHE IVa, as the sole predictor of in-hospital mortality and fitted a logistic regression model trained on the entire pooled dataset. The fitted model was then tested separately for each target diagnosis group. %\SM{Add a reference in regards to the Apache score being a good predictor. }

\paragraph{Target-only models.}
We also evaluated models trained solely within each diagnostic group, using all baseline covariates as predictors for the XGBoost models and including pairwise interaction terms only for the logistic regression models, since XGBoost inherently models nonlinearities and feature interactions without requiring explicit interaction terms. These two target-only models underwent the same recalibration procedure described in \autoref{Recalibration} to ensure a fair comparison of calibration across methods. 

\paragraph{Source models.}
We trained XGBoost and GLM models using the entire source dataset, including all baseline covariates. For the GLM based source models, we also include all pairwise interactions. To improve calibration, we applied the same recalibration procedure as described in \autoref{Recalibration}, using the source data. The calibrated model was then directly applied to each target diagnostic group without further adaptation.

\subsection{Model evaluation}
We evaluated predictive model performance in terms of discrimination using the area under the receiver operating characteristic curve (AUROC) and the area under the precision–recall curve (AUPRC), calibration using the Integrated Calibration Index (ICI) \citep{austin2019integrated}, and an overall joint measure of both discrimination and calibration using the Brier score.

AUPRC summarizes model performance by integrating precision as a function of recall across all classification thresholds and is particularly informative in imbalanced outcome settings, such as low-prevalence mortality outcomes (noting that in-hospital mortality is 8.72\% in our eICU-CRD cohort and as low as 1.69\% among the diagnosis groups).
$\text{Precision} = \frac{\text{TP}}{\text{TP} + \text{FP}}$ represents the proportion of predicted positives that are truly positive, while
$\text{recall (or sensitivity)} = \frac{\text{TP}}{\text{TP} + \text{FN}}$ represents the proportion of actual positives that the model correctly predicts.

In addition, we assessed the predictive accuracy under different decision thresholds through sensitivity, specificity, balanced accuracy (\autoref{eq:balance_acc}), and F1 score (\autoref{eq:f1_score}).

\begin{equation}
    \text{Balanced accuracy} = \frac{1}{2} \times(\text{sensitivity}+ \text{specificity})
    \label{eq:balance_acc}
\end{equation}

\begin{equation}
    \text{F1 score} = 2 \times \frac{\text{Precision} \times \text{Recall}}{\text{Precision} + \text{Recall}}
    \label{eq:f1_score}
\end{equation}
% predictive accuracy

\section{Data application}\label{ApplicationResult}
For illustration, we selected four diagnostic groups from eICU-CRD here -- ARF, CardiacArrest, Sepsis, and CVA -- representing, respectively, small total sample size, the highest prevalence, the largest total sample size, and the second largest total sample size (\autoref{tab:diagnosis_summary}). 
% Results for all diagnostic groups and all methods are provided in the Supplementary Materials. 

\subsection{Discrimination and calibration performance}
% Add a table summarizing performance of apache only, target only, source models and transfer learnings, in order to illustrate the limited performance of apache/target only models. 
\begin{table}[ht]
\centering
\caption{AUROC and AUPRC results by four diagnoses and seven methods. Values represent mean (standard deviation) of AUROC and AUPRC from all test folds. "TL XGB" and "TL GLM" denote transfer learning models for XGB and GLM respectively. }
\label{Table2}
\resizebox{\textwidth}{!}{%
\begin{tabular}{lccccccc}
\toprule
\multicolumn{8}{c}{\textbf{AUROC:} Mean(SD)} \\
\midrule
Diagnosis & APACHE only & Target only GLM & Source GLM & TL GLM & Target only XGB & Source XGB & TL XGB \\
\midrule
ARF            & 0.76 (0.07) & 0.81 (0.06) & 0.84 (0.05) & 0.84 (0.05) & 0.83 (0.03) & 0.86 (0.03) & 0.86 (0.03) \\
CVA            & 0.83 (0.02) & 0.88 (0.01) & 0.89 (0.02) & 0.89 (0.02) & 0.90 (0.01) & 0.90 (0.02) & 0.90 (0.02) \\
CardiacArrest  & 0.88 (0.01) & 0.90 (0.00) & 0.91 (0.01) & 0.91 (0.01) & 0.92 (0.01) & 0.92 (0.01) & 0.92 (0.01) \\
Sepsis         & 0.75 (0.01) & 0.83 (0.01) & 0.83 (0.01) & 0.84 (0.01) & 0.85 (0.01) & 0.84 (0.01) & 0.85 (0.01) \\
\midrule
\multicolumn{8}{c}{\textbf{AUPRC:} Mean(SD)} \\
\midrule
Diagnosis & APACHE only & Target only GLM & Source GLM & TL GLM & Target only XGB & Source XGB & TL XGB \\
\midrule
ARF            & 0.35 (0.11) & 0.41 (0.11) & 0.44 (0.10) & 0.35 (0.21) & 0.40 (0.11) & 0.52 (0.08) & 0.51 (0.07) \\
CVA            & 0.44 (0.05) & 0.49 (0.05) & 0.52 (0.05) & 0.53 (0.05) & 0.57 (0.03) & 0.59 (0.05) & 0.59 (0.05) \\
CardiacArrest  & 0.68 (0.01) & 0.73 (0.01) & 0.73 (0.01) & 0.73 (0.01) & 0.77 (0.01) & 0.78 (0.01) & 0.78 (0.01) \\
Sepsis         & 0.40 (0.02) & 0.57 (0.02) & 0.56 (0.02) & 0.58 (0.01) & 0.61 (0.02) & 0.61 (0.01) & 0.61 (0.01) \\
\bottomrule
\end{tabular}%
}
\end{table}

First, we evaluate the AUROC and AUPRC scores of seven models: the Target-only models, the Source models, and the Transfer Learning models (adjusting for main effects) for both GLM and XGBoost, as well as the APACHE-only model. The summarized mean and standard deviation of AUROC and AUPRC across diagnoses and methods are shown in \autoref{Table2}.

Across all four diagnoses, the APACHE-only model consistently demonstrated the poorest performance in both AUROC and AUPRC, suggesting that relying solely on APACHE scores is inadequate for mortality risk prediction. Likewise, the Target-only models generally exhibited poorer performance compared to the source and transfer learning models, highlighting the limitation of using only the restricted target data in terms of model discrimination. 
An exception is observed in CVA, where the Target-only GLM achieved a higher AUPRC than the source model and performed comparably to both the source and transfer learning models in terms of AUROC. This is partially due to the relatively large sample size of the CVA cohort, which provides enough information for robust training within the target domain. On the other hand, the data distribution of CVA is more distinct from that of other diagnoses, so source models tend to be biased when applied to CVA patients.

Overall, the XGBoost-based models consistently outperform the GLM-based models in terms of AUROC and AUPRC. Moreover, as shown in \autoref{Table2}, the GLM-based transfer learning model yields marginal improvements on AUROC and AUPRC over the GLM-based source model in certain diagnoses, whereas the XGBoost-based transfer learning model performs almost identically in AUROC and AUPRC to the XGBoost source model.

\begin{figure}
    \centering
    \includegraphics[width=0.9\linewidth]{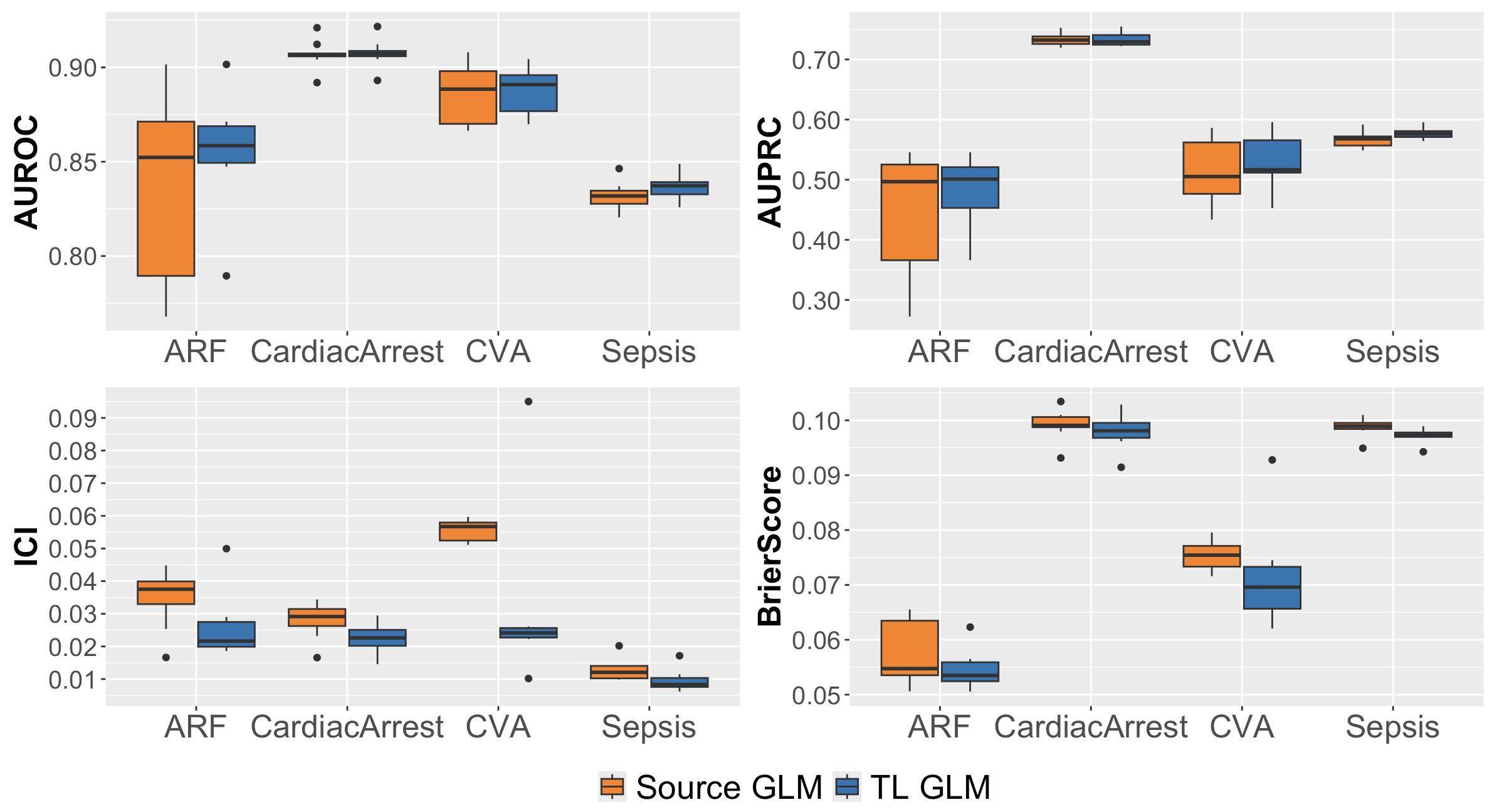}
    \caption{GLM Discrimination and calibration performance for source and transfer learning models. Metric values are obtained from all test folds. X-axis represents diagnoses.}
    \label{fig:GLM_DC}
\end{figure}

\begin{figure}
    \centering
    \includegraphics[width=0.9\linewidth]{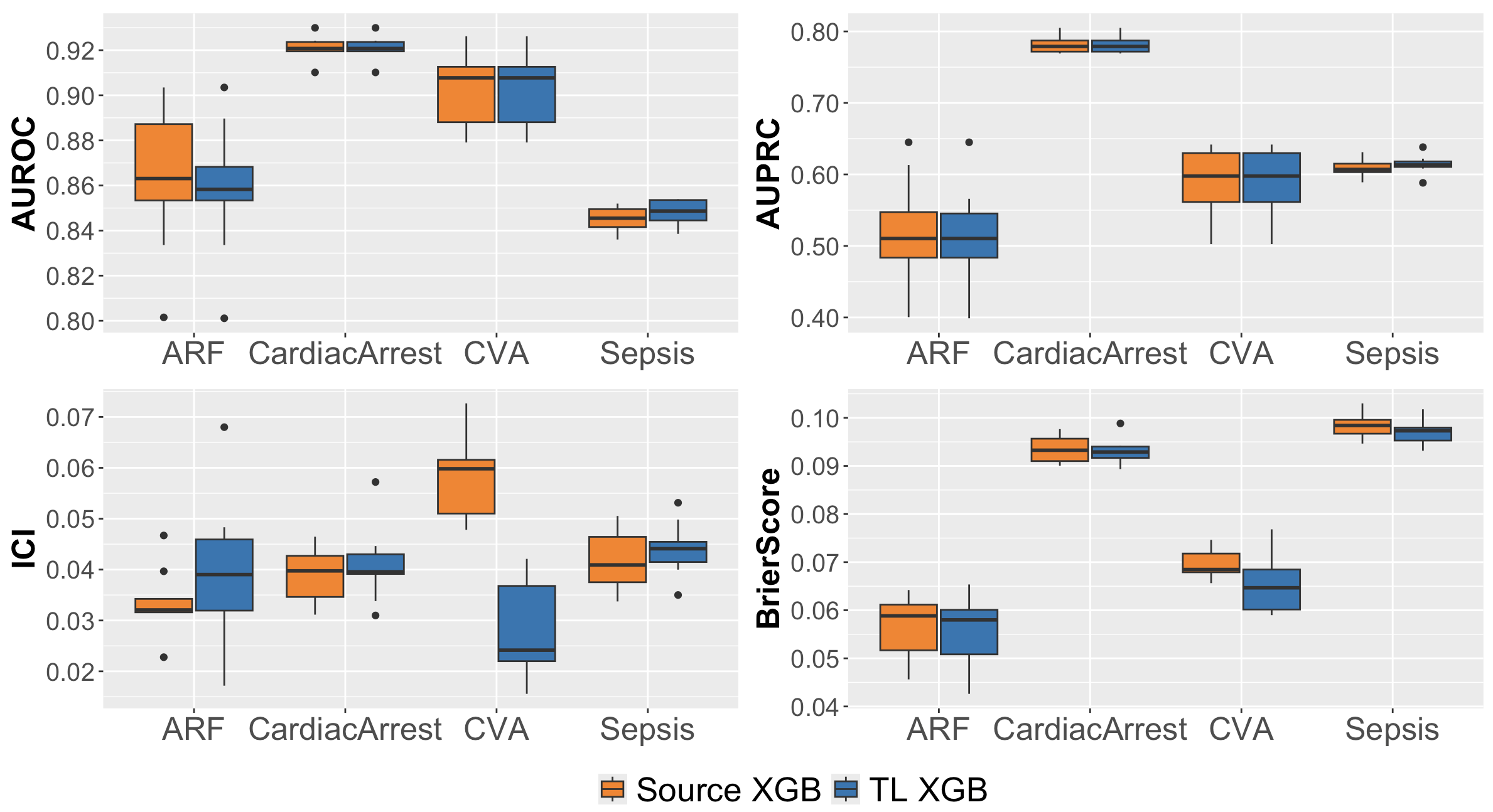}
    \caption{XGBoost discrimination and calibration performance for source and transfer learning models. Metric values are obtained from all test folds. X-axis represents diagnoses.}
    \label{fig:XGB_DC}
\end{figure}

%Then include source & TL models due to insufficient information in the plots to compare their performance:
To further explore the improvement of transfer learning on source models 
% in the eICU-CRD 
on discrimination and calibration performances, we compared the AUROC, AUPRC, ICI, and Brier Score of the transfer learning models and source models for both GLM (\autoref{fig:GLM_DC}) and XGBoost (\autoref{fig:XGB_DC}). 

Transfer learning yielded modest improvements in AUROC and AUPRC over the source GLM for Sepsis, while achieved similar AUROC and AUPRC scores with the source models. The lack of large gains in discrimination metrics can be attributed to a possible ceiling effect — the source models have already achieved very high AUROC and AUPRC, leaving limited room for further improvement.
In \autoref{simulation}, further investigation of such a ceiling effect is pursued through simulation studies.

When examining calibration, transfer learning methods generally reduced the Integrated Calibration Index (ICI) and the Brier score compared to the source models, especially in GLM-based models. This indicates that, regardless of the modeling framework, transfer learning enhanced the alignment between predicted probabilities and observed outcomes, resulting in more reliable risk estimation.

\subsection{Cutoff threshold and predictive accuracy}
\begin{figure}[htbp]
    \centering
    \includegraphics[width=1\linewidth]{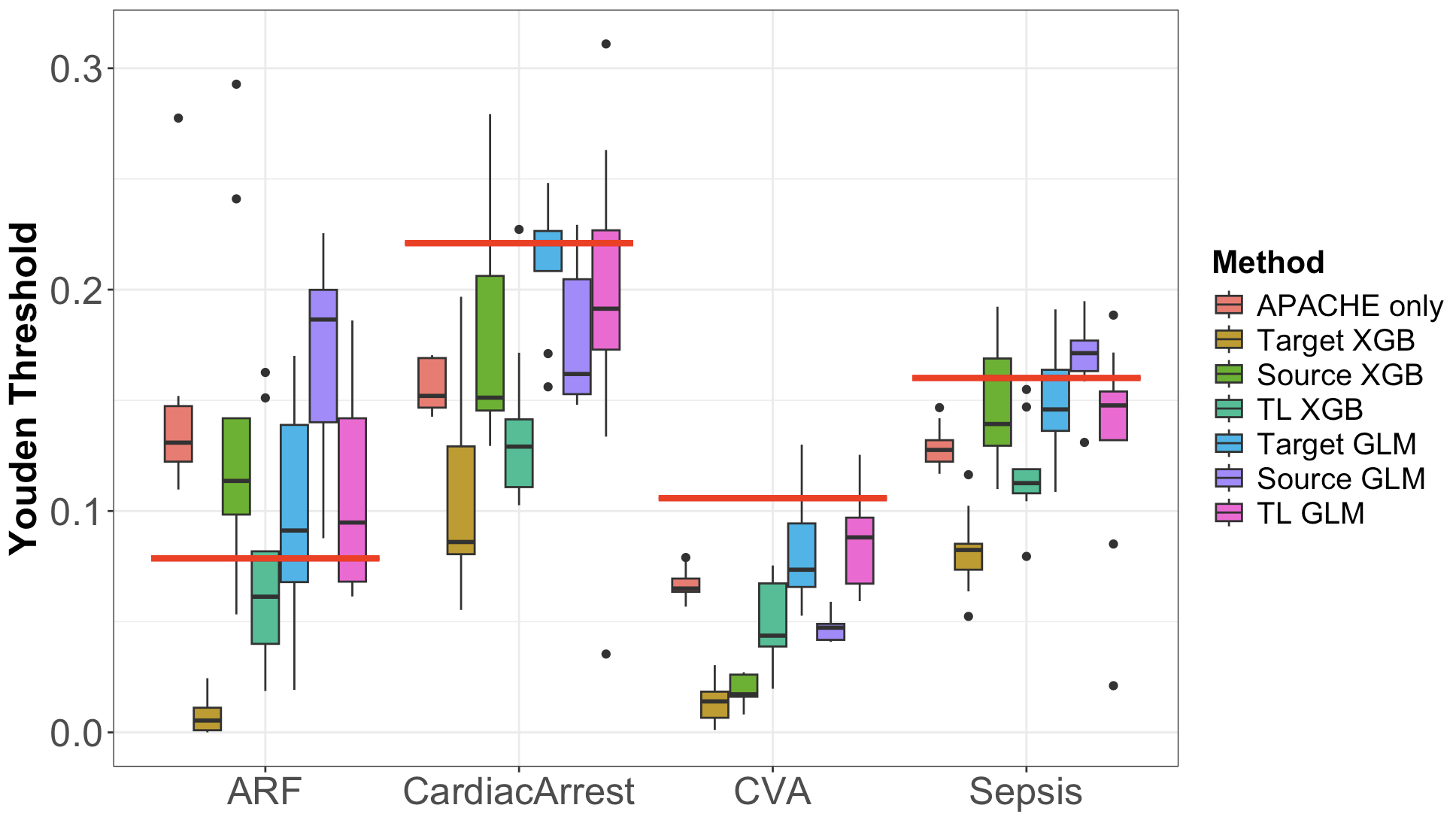}
    \caption{Youden cutoff (boxes) v.s. prevalence (red horizontal lines) as decision thresholds. The Youden cutoff is calculated from different training folds across diagnoses and methods. The prevalence is the mortality rate per diagnosis. X-axis represents different diagnoses, and y-axis represents decision threshold values.}
    \label{fig:cutoff}
\end{figure}

Since the prevalence of in-hospital mortality in eICU-CRD is very low for most diagnostic groups (\autoref{tab:diagnosis_summary}), we explored alternative classification thresholds beyond the conventional 0.5 cutoff. 
Specifically, we examined thresholds based on the Youden index -- defined as the point on the ROC curve that maximizes $\text{sensitivity} + \text{specificity} - 1$, and the observed prevalence for each diagnosis. 
\autoref{fig:cutoff} compares the Youden cutoff with the prevalence across all diagnostic groups and modeling approaches. The plot shows that the Youden threshold is reasonably consistent with the prevalence across diagnostic groups. 

\autoref{fig:predacc} summarizes the predictive accuracy of transfer learning models compared with source models across the three decision thresholds. 
Under all cutoffs, transfer learning mostly achieves higher sensitivity, balanced accuracy, and F1 scores than source models, indicating robust performance in diagnostically heterogeneous groups. 

In addition, applying the conventional 0.5 cutoff generally yields the lowest sensitivity and balanced accuracy. Despite achieving high F1 scores, the regular cutoff of 0.5 produces very low sensitivity. The conservative threshold causes the classifier to identify only a small subset of positives, which greatly inflates precision but misses many true cases. 
In contrast, the Youden cutoff consistently achieved the highest balanced accuracy, which is expected given the low prevalence of in-hospital mortality and the property of the Youden index to maximize balanced accuracy.

\begin{table}[h]
\centering
\caption{Predictive accuracy of GLM and XGBoost models for ARF, CardiacArrest, CVA and Sepsis under three classification thresholds: regular 0.5 cutoff, Youden index, and prevalence. Entries are mean (standard deviation) from all test folds. “Source” = source model; “TL” = transfer learning model. \textbf{Bold indicates TL $>$ Source.}}
\label{fig:predacc}
\renewcommand{\arraystretch}{1.05}
\setlength{\tabcolsep}{6pt}
\resizebox{\textwidth}{!}{%
\begin{tabular}{ll|cc|cc|cc}
\toprule
\multicolumn{2}{c}{} &
\multicolumn{2}{c}{\textbf{Sensitivity}} &
\multicolumn{2}{c}{\textbf{Balanced accuracy}} &
\multicolumn{2}{c}{\textbf{F1}} \\
\textbf{Threshold} & \textbf{Target} &
\textbf{Source} & \textbf{TL} &
\textbf{Source} & \textbf{TL} &
\textbf{Source} & \textbf{TL} \\
\midrule
\multicolumn{8}{l}{\textbf{GLM}} \\
\addlinespace[2pt]

% ===== GLM 0.5 =====
\multirow{4}{*}{0.5}
  & ARF           & 0.276 (0.087) & 0.235 (0.093) & 0.629 (0.041) & 0.612 (0.044) & 0.961 (0.004) & \textbf{0.963 (0.004)} \\
  & CardiacArrest & 0.590 (0.066) & \textbf{0.609 (0.035)} & 0.759 (0.026) & \textbf{0.766 (0.015)} & 0.908 (0.003) & 0.908 (0.003) \\
  & CVA           & 0.209 (0.12)  & \textbf{0.302 (0.093)} & 0.596 (0.053) & \textbf{0.639 (0.044)} & 0.947 (0.003) & \textbf{0.948 (0.004)} \\
  & Sepsis        & 0.353 (0.03)  & 0.340 (0.044) & 0.661 (0.013) & 0.656 (0.019) & 0.926 (0.002) & \textbf{0.927 (0.002)} \\

\addlinespace[2pt]\midrule

% ===== GLM Prevalence =====
\multirow{4}{*}{Prevalence}
  & ARF           & 0.731 (0.134) & 0.731 (0.097) & 0.740 (0.048) & \textbf{0.760 (0.041)} & 0.842 (0.063) & \textbf{0.871 (0.018)} \\
  & CardiacArrest & 0.845 (0.029) & \textbf{0.858 (0.026)} & 0.840 (0.010) & \textbf{0.842 (0.009)} & 0.888 (0.005) & 0.885 (0.006) \\
  & CVA           & 0.710 (0.098) & \textbf{0.751 (0.194)} & 0.788 (0.032) & \textbf{0.792 (0.073)} & 0.911 (0.020) & 0.894 (0.022) \\
  & Sepsis        & 0.727 (0.028) & \textbf{0.729 (0.038)} & 0.752 (0.009) & \textbf{0.755 (0.009)} & 0.849 (0.013) & \textbf{0.852 (0.014)} \\

\addlinespace[2pt]\midrule

% ===== GLM Youden =====
\multirow{4}{*}{Youden}
  & ARF           & 0.707 (0.089) & \textbf{0.715 (0.105)} & 0.781 (0.048) & \textbf{0.791 (0.045)} & 0.907 (0.048) & \textbf{0.916 (0.024)} \\
  & CardiacArrest & 0.874 (0.031) & 0.869 (0.032) & 0.846 (0.008) & \textbf{0.847 (0.009)} & 0.882 (0.009) & \textbf{0.886 (0.013)} \\
  & CVA           & 0.853 (0.026) & \textbf{0.856 (0.037)} & 0.821 (0.015) & \textbf{0.823 (0.014)} & 0.873 (0.015) & \textbf{0.874 (0.017)} \\
  & Sepsis        & 0.738 (0.035) & \textbf{0.761 (0.032)} & 0.756 (0.008) & \textbf{0.761 (0.007)} & 0.848 (0.020) & 0.842 (0.021) \\

\midrule
\multicolumn{8}{l}{\textbf{XGBoost}} \\
\addlinespace[2pt]

% ===== XGB 0.5 =====
\multirow{4}{*}{0.5}
  & ARF           & 0.300 (0.148) & \textbf{0.391 (0.076)} & 0.641 (0.071) & \textbf{0.684 (0.038)} & 0.962 (0.005) & \textbf{0.963 (0.006)} \\
  & CardiacArrest & 0.606 (0.140) & \textbf{0.680 (0.032)} & 0.767 (0.061) & \textbf{0.802 (0.011)} & 0.910 (0.008) & \textbf{0.917 (0.003)} \\
  & CVA           & 0.271 (0.179) & \textbf{0.404 (0.031)} & 0.626 (0.081) & \textbf{0.688 (0.017)} & 0.949 (0.004) & \textbf{0.952 (0.005)} \\
  & Sepsis        & 0.477 (0.015) & 0.455 (0.017) & 0.710 (0.007) & 0.703 (0.008) & 0.923 (0.003) & \textbf{0.926 (0.003)} \\

\addlinespace[2pt]\midrule

% ===== XGB Prevalence =====
\multirow{4}{*}{Prevalence}
  & ARF           & 0.800 (0.083) & 0.687 (0.098) & 0.712 (0.071) & \textbf{0.769 (0.045)} & 0.754 (0.100) & \textbf{0.906 (0.011)} \\
  & CardiacArrest & 0.805 (0.057) & \textbf{0.828 (0.032)} & 0.830 (0.023) & \textbf{0.843 (0.010)} & 0.895 (0.005) & \textbf{0.900 (0.005)} \\
  & CVA           & 0.628 (0.062) & \textbf{0.757 (0.033)} & 0.768 (0.028) & \textbf{0.809 (0.017)} & 0.930 (0.012) & 0.912 (0.009) \\
  & Sepsis        & 0.695 (0.047) & \textbf{0.703 (0.017)} & 0.757 (0.010) & \textbf{0.761 (0.008)} & 0.872 (0.019) & \textbf{0.874 (0.008)} \\

\addlinespace[2pt]\midrule

% ===== XGB Youden =====
\multirow{4}{*}{Youden}
  & ARF           & 0.733 (0.121) & \textbf{0.791 (0.083)} & 0.766 (0.056) & \textbf{0.801 (0.036)} & 0.874 (0.055) & \textbf{0.885 (0.050)} \\
  & CardiacArrest & 0.868 (0.039) & \textbf{0.886 (0.036)} & 0.844 (0.017) & \textbf{0.854 (0.011)} & 0.883 (0.009) & \textbf{0.887 (0.008)} \\
  & CVA           & 0.846 (0.059) & \textbf{0.885 (0.025)} & 0.815 (0.030) & \textbf{0.834 (0.016)} & 0.869 (0.022) & \textbf{0.871 (0.026)} \\
  & Sepsis        & 0.754 (0.027) & \textbf{0.768 (0.037)} & 0.766 (0.008) & \textbf{0.768 (0.008)} & 0.853 (0.018) & 0.847 (0.021) \\

\bottomrule
\end{tabular}
}
\end{table}

\section{Simulation}\label{simulation}
In the eICU dataset, due to the ceiling effect of AUROC scores, it is challenging to observe meaningful improvements in discrimination performance when comparing transfer learning models to source-only models. To better illustrate the advantages of transfer learning, we therefore conducted a simulation study in which the source model exhibits moderate AUROC performance, allowing greater room for improvement and a clearer demonstration of transfer learning benefits.

\subsection{Data generation}
To mimic the real-data setting, we design a simplified simulation study focusing on a single target dataset ($K=1$). A total of $n=2000$ observations are generated, where the target dataset has size $m=200$ and the remaining $n-m=1800$ samples are used as source data. The source model is trained on the combined dataset of size $n=2000$.

To control both prevalence and signal strength, we adopt a class-conditional Gaussian data generating mechanism. Specifically, for each observation $i \in 1, \ldots, n$, the binary outcome $Y_i \in \{-1,1\}$ is generated from a Rademacher distribution,
\[Y_i \sim \text{Rad}(\pi),\] 
where $\pi$ denotes the prevalence. Conditional on $Y_i$, the $p=10$ covariates $\boldsymbol{X}_i \in \mathbb{R}^{p}$ are generated from
\[
\boldsymbol{X}_i|Y_i \stackrel{iid}{\sim} N(\beta Y_i, \boldsymbol{\Sigma}_p),
\]
where 
\[
\beta =
\begin{cases}
(a, a, b, b, b, 0, 0, 0, 0, 0), & \text{for target data}, \\
(-a, -a, b, b, b, 0, 0, 0, 0, 0), & \text{for source data},
\end{cases}
\]
represents the correlation between $Y_i$ and $\boldsymbol{X}_i$, and 
$
\boldsymbol{\Sigma}_p = diag(\boldsymbol\Sigma_{2}, \boldsymbol I_{p-2} )
$ is the covariance matrix, 
where 
\[
\boldsymbol\Sigma_2 = 
\begin{cases}
    \begin{pmatrix}
    1 & 0 \\
    0 & 1
    \end{pmatrix}, Y_i = 1 \\\\
    
    \begin{pmatrix}
    3 & 0.4 \\
    0.4 & 3
    \end{pmatrix}, Y_i = -1
\end{cases}
\]
This construction of $\boldsymbol\Sigma_p$ introduces both linear and non-linear components into the data generation process, in order to better mimic realistic data settings.

Note that the heterogeneity of $Y|\boldsymbol{X}$ between source and target is controlled by the value of $a$, and the discrimination of the data is controlled by the the value of $a$ and $b$: Small values of $b$ reduce the discrimination and small values of $a$ reduce both the heterogeneity and the discrimination. 

To mimic the low prevalence of mortality in the real eICU data, we consider a setting with $\pi = 0.1$. In addition, to examine a more general case with balanced outcomes, we include $\pi = 0.5$ as a standard scenario. We set $a = 0.3$ to control the degree of heterogeneity, and evaluate $b \in \{0.2, 0.7\}$ under each prevalence level, which generates four scenarios in total. This design allows us to evaluate our methods under varying levels of both outcome prevalence and discrimination with controlled source–target heterogeneity. The simulation repeats 20 times in each scenario.

\subsection{Result}
\begin{figure}[h]
    \centering
    \begin{subfigure}{0.45\textwidth}
        \centering
        \includegraphics[width=\linewidth]{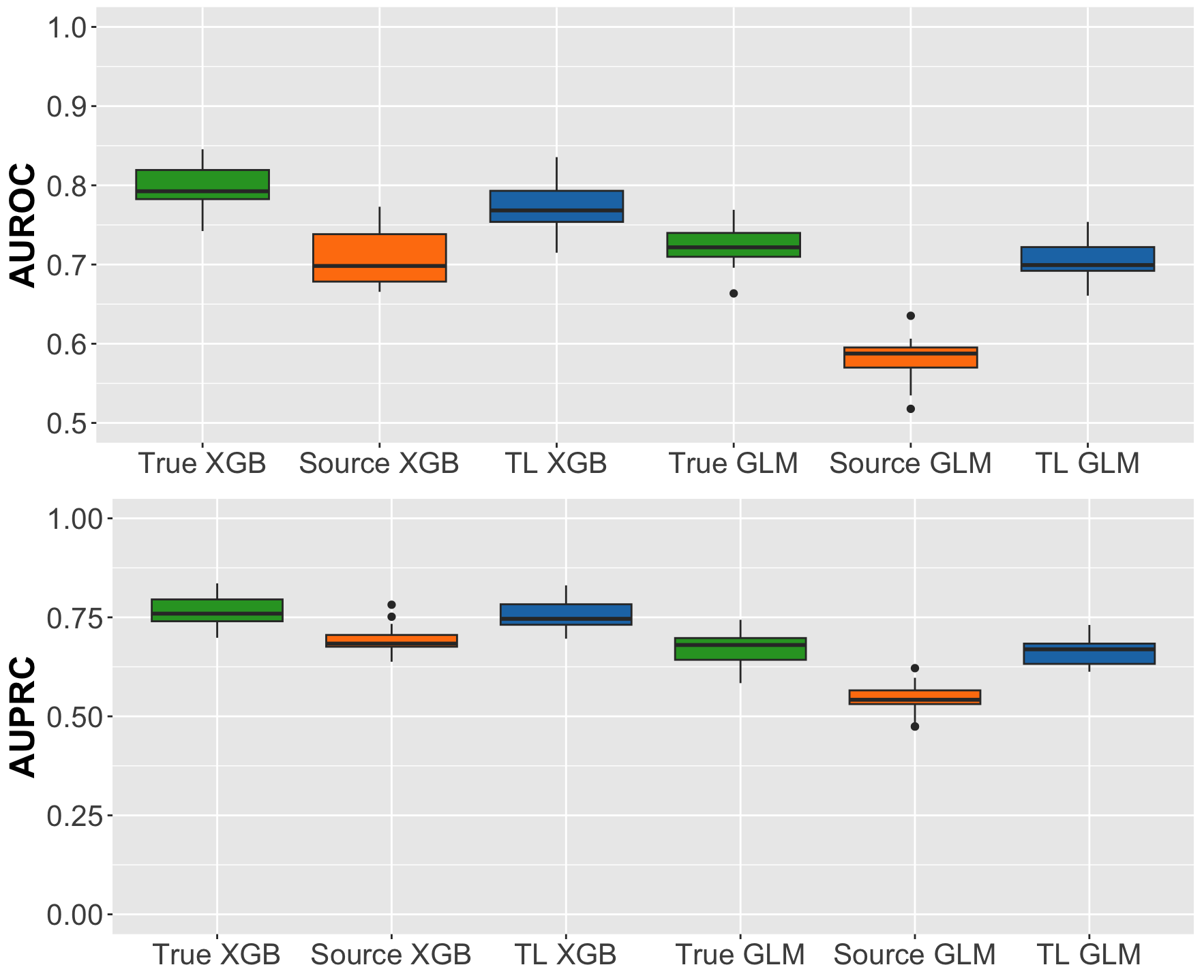}
        \caption{prevalence $ = 0.5, \ b = 0.2$}
        \label{fig:2x2a}
    \end{subfigure}
    \begin{subfigure}{0.45\textwidth}
        \centering
        \includegraphics[width=\linewidth]{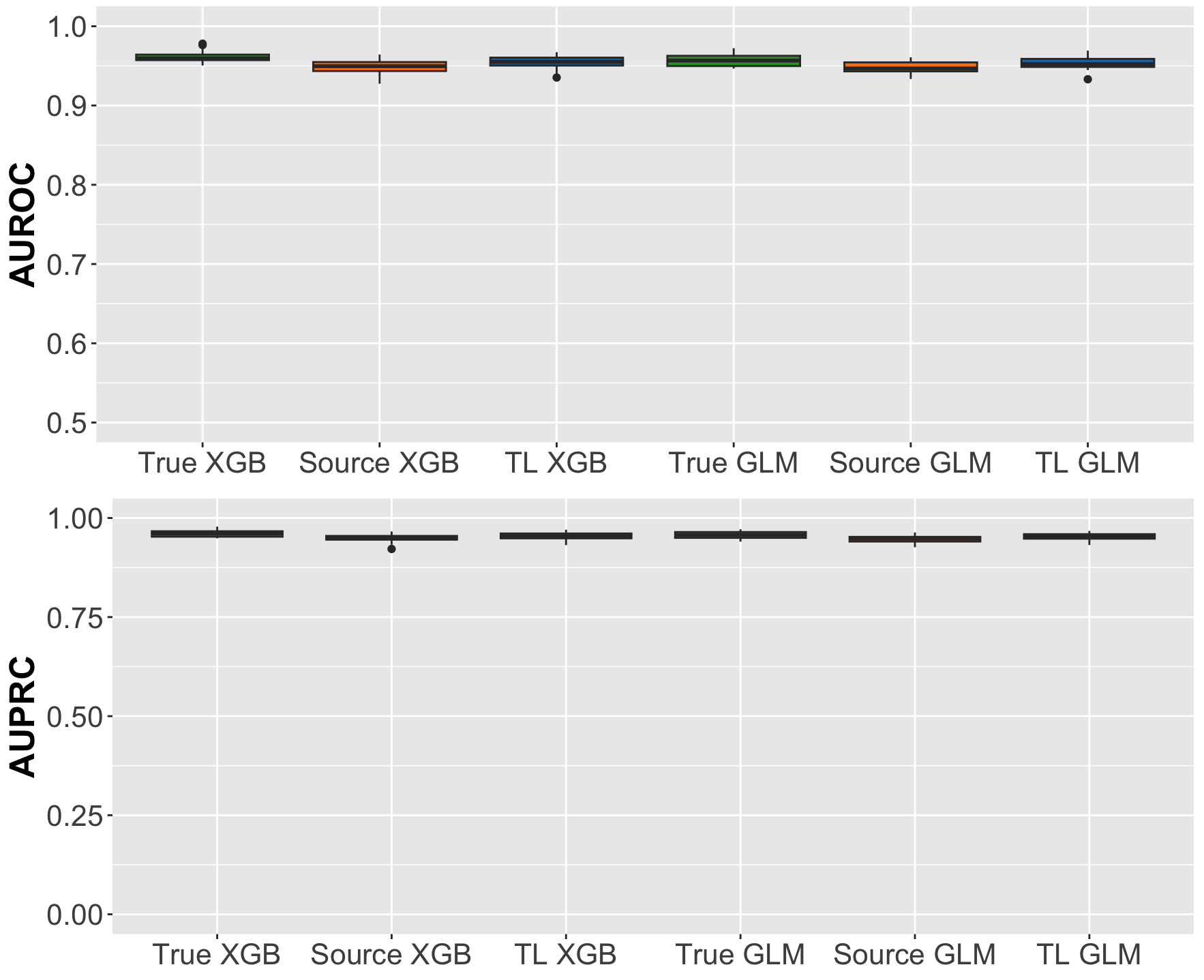}
        \caption{prevalence $ = 0.5, \ b = 0.7$}
        \label{fig:2x2b}
    \end{subfigure}
    
    \vspace{0.5cm} 

    \begin{subfigure}{0.45\textwidth}
        \centering
        \includegraphics[width=\linewidth]{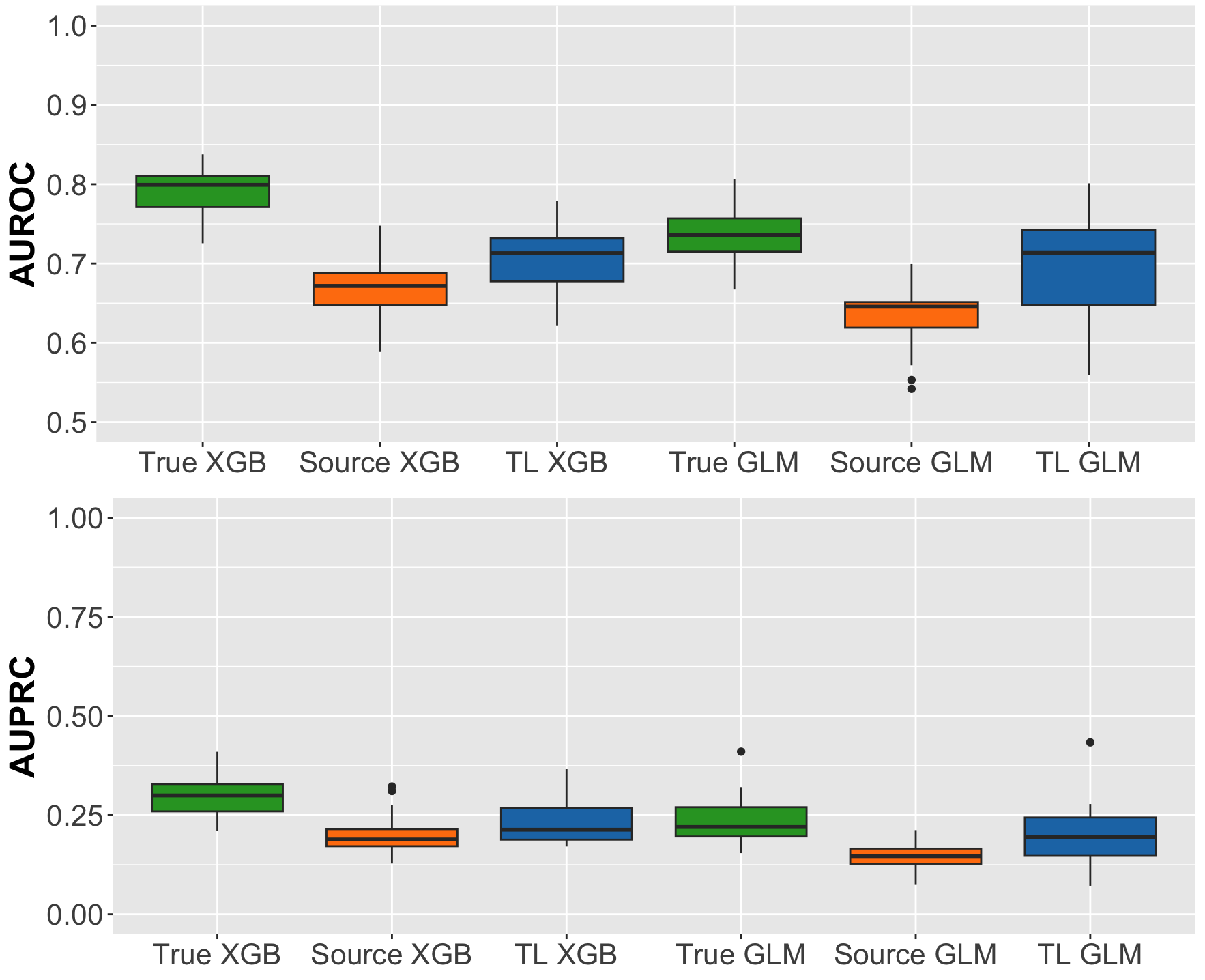}
        \caption{prevalence $ = 0.1, \ b = 0.2$}
        \label{fig:2x2c}
    \end{subfigure}
    \begin{subfigure}{0.45\textwidth}
        \centering
        \includegraphics[width=\linewidth]{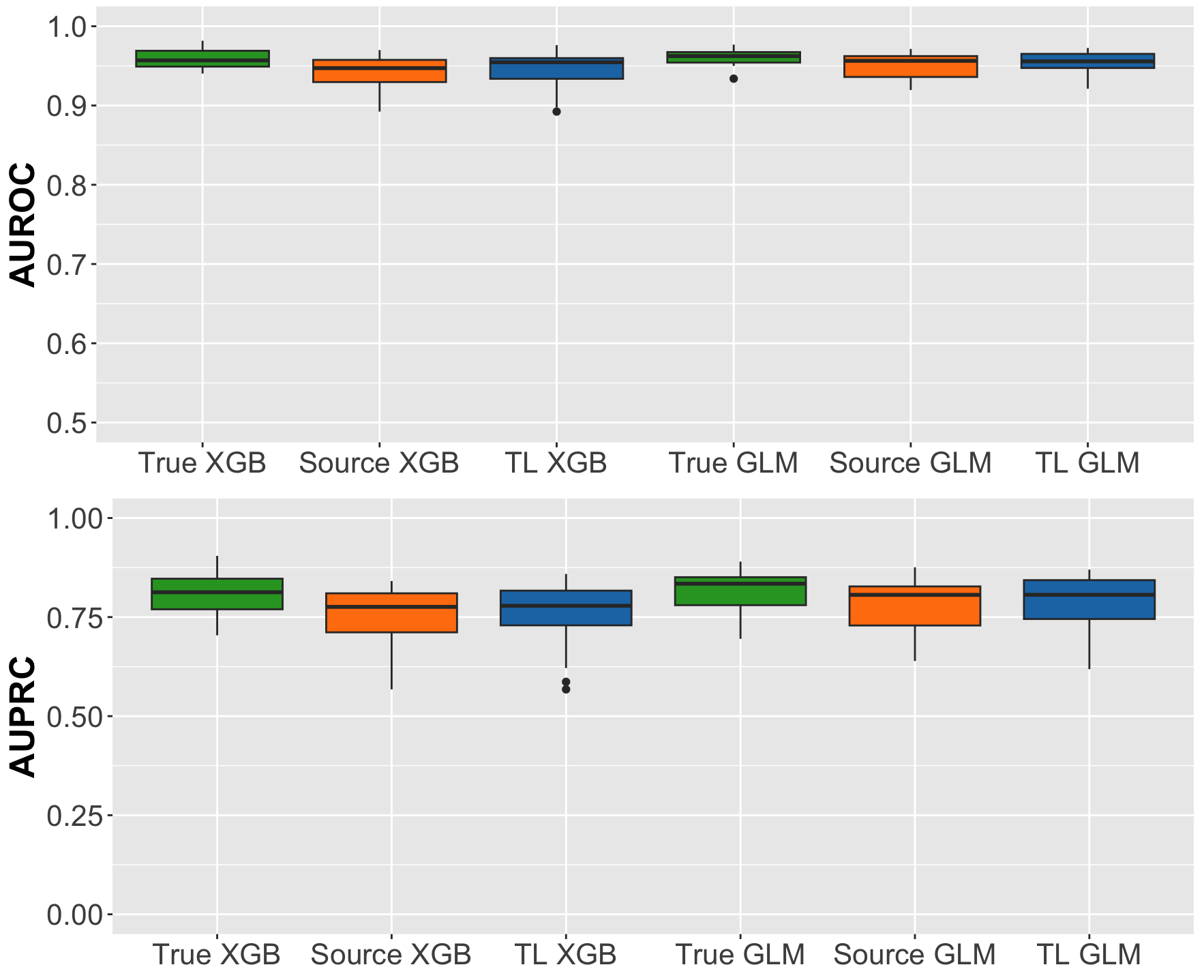}
        \caption{prevalence $ = 0.1, \ b = 0.7$}
        \label{fig:2x2d}
    \end{subfigure}
    
    \caption{Simulation results of AUROC and AUPRC, comparing true models, source models and transfer learning models for XGBoost and GLM under four scenarios. $b$ is signal strength, representing discrimination level. X-axis represents different methods.}
    \label{fig:2x2}
\end{figure}

\autoref{fig:2x2} summarizes the simulation results in terms of AUROC and AUPRC under the four scenarios. 
Overall, AUPRC scores decline under lower prevalence ($\pi = 0.1$) relative to moderate prevalence ($\pi = 0.5$) across all models, reflecting the impact of class imbalance. When prevalence is lower, the improvements in AUROC and AUPRC from transfer learning over source models become somewhat smaller than under moderate prevalence. Nevertheless, at both prevalence levels, when the signal strength is modest ($b=0.2$), transfer learning models still achieve clear improvements in both AUROC and AUPRC compared with models trained only on the source data (\autoref{fig:2x2a}, \autoref{fig:2x2c}). As the signal strength increases ($b=0.7$), the source models already attain high AUROC and AUPRC values close to those of the true models, leaving little room for transfer learning to provide further gains (\autoref{fig:2x2b}, \autoref{fig:2x2d}). This ceiling effect is consistent with what we observe in the real eICU data.

% low prevalence for mimic the real data; also put 0.5 preval here for standard; more in supplement.

\section{Discussion}\label{Discussion}
In our study, we investigated both GLM- and XGBoost-based transfer learning approaches for predicting mortality among ICU patients across different diagnostic groups, an area of significant clinical relevance. While GLMs generally underperform XGBoost in accuracy, they remain valuable for their interpretability, and transfer learning enhances their performance without sacrificing this advantage. 
As a case study, we applied these methods to the eICU-CRD database to predict in-hospital mortality across diagnostic groups. In the eICU-CRD, transfer learning often achieves better calibration than source models, underscoring its clinical utility, as reliable probabilities are crucial for decision-making. In addition, we observed a ceiling effect in settings with strong baseline discrimination; however, simulations suggest that transfer learning can yield substantial gains where baseline models perform less effectively.
Finally, given the low mortality prevalence in eICU-CRD, conventional 0.5 thresholds were suboptimal; nevertheless, across different thresholds, transfer learning delivered robust and stable improvements in both GLM- and XGBoost-based models, highlighting its broad utility in heterogeneous ICU populations.

Our transfer learning models are not limited to predicting ICU mortality. When paired with routinely collected electronic health records, they can support emergency department triage, deterioration risk prediction, and generalize to other clinically relevant endpoints (e.g., hospital length of stay, 30-day readmission, and laboratory/vital-sign trajectories). Future work includes extending this framework to external validation across untrained regions, institutions, or demographic subgroups. Such efforts will be essential to establish the generalizability of transfer learning for ICU mortality prediction and to confirm its value in diverse real-world clinical settings.

\bigskip

%\break
\bigskip 

 % \newcommand{\noop}[1]{}

% \printbibliography
\bibliographystyle{unsrtnat}
\bibliography{ref}
\end{document}